\RequirePackage[2022/05/31]{latexrelease}
\documentclass[letterpaper, 10 pt, conference]{ieeeconf}  % Comment this line out if you need a4paper

\IEEEoverridecommandlockouts                              % This command is only needed if 
\usepackage{graphicx}
\usepackage{algorithm}
\usepackage{algorithmicx}
\usepackage{algpseudocode}
\usepackage{amsmath}
\usepackage{amsfonts}
\usepackage{subcaption}
\usepackage{nidanfloat}

\title{\LARGE \bf
Autonomous Cooperative Transportation System \\
involving Multi-Aerial Robots with Variable Attachment Mechanism
}

\author{Koshi Oishi$^{1}$ and Tomohiko Jimbo$^{1}$% <-this % stops a space
\thanks{$^{1}$The authors are with the Cloud Informatics Research-Domain,
        Toyota Central R\&D Labs., Inc., 41-1, Yokomichi, Nagakute, Aichi, Japan
        {\tt\small e1616@mosk.tytlabs.co.jp}}%
}

\begin{document}

\maketitle
\thispagestyle{empty}
\pagestyle{empty}

\begingroup
  \renewcommand\thefootnote{}            % 脚注マークを出さない
  \footnotetext{\footnotesize
  © 2025 IEEE. Personal use of this material is permitted. 
  Permission from IEEE must be obtained for all other uses, in any current or future media, 
  including reprinting/republishing this material for advertising or promotional purposes, 
  creating new collective works, for resale or redistribution to servers or lists, 
  or reuse of any copyrighted component of this work in other works.}
  \addtocounter{footnote}{-1}            % カウンタを戻す
\endgroup

% max 200 word
%%%%%%%%%%%%%%%%%%%%%%%%%%%%%%%%%%%%%%%%%%%%%%%%%%%%%%%%%%%%%%%%%%%%%%%%%%%%%%%%
\begin{abstract}
    Cooperative transportation by multi-aerial robots has the potential to support various payloads and improve fail-safe against dropping.
    Furthermore, changing the attachment positions of robots according payload characteristics increases the stability of transportation.
    However, there are almost no transportation systems capable of scaling to the payload weight and size and changing the optimal attachment positions.
    To address this issue, we propose a cooperative transportation system comprising autonomously executable software and suitable hardware,
    covering the entire process, from pre-takeoff setting to controlled flight.
    The proposed system decides the formation of the attachment positions by prioritizing controllability
    based on the center of gravity obtained from Bayesian estimations with robot pairs.
    We investigated the cooperative transportation of an unknown payload larger than that of whole carrier robots through numerical simulations.
        Furthermore, we performed cooperative transportation of an unknown payload (with a weight of about $3.2$ kg and maximum length of $1.76$ m) using eight robots.
    The proposed system was found to be versatile with regard to handling unknown payloads with different shapes and center-of-gravity positions.

\end{abstract}

% \begin{IEEEkeywords}
% Intelligent Transportation Systems
% Cooperating Robots
% Multi-Robot Systems
% Swarm Robotics

% \end{IEEEkeywords}

%%%%%%%%%%%%%%%%%%%%%%%%%%%%%%%%%%%%%%%%%%%%%%%%%%%%%%%%%%%%%%%%%%%%%%%%%%%%%%%%
\section{INTRODUCTION}
Recently, aerial robots have been used in various fields such as photography, agriculture, and transportation \cite{c1,c2,c2_1}.
In particular, as aerial robots capable of three-dimensional movement do not require the construction of ground infrastructure, 
transportation by aerial robots has entered the practical stage in a limited environment. 
However, a major setback is the limited payload of a single aerial robot. 
With the use of only one aerial robot, 
preliminary preparations are required to ensure that the center-of-gravity (COM) and mass (physical parameters) of the payload satisfy the specifications of the aerial robot.
This limitation can be overcome by implementing cooperative transportation using multiple aerial robots. 
By employing multiple aerial robots, 
payload limitations are expected to be relaxed in terms of the shape and mass, and fail-safe against dropping is also expected to be improved. 
With this approach, only the required number of robots would need to be gathered.
 Therefore, cooperative transportation by multiple aerial robots is attracting attention \cite{c3,c3_05}.
In addition, if the aerial robots can estimate the physical parameters of the payload by themselves and determine the optimal attachment positions, 
an aerial transport service without human intervention can be achieved.

Cooperative transportation by aerial robots has been extensively studied through modeling \cite{c3_1,c4,c5}, leader follower control \cite{c6}, and distributed control \cite{c7,c8}.
In cooperative transportation, even takeoff is difficult if the attachment positions are not appropriate for the physical parameters. 
Therefore, in many existing studies, the physical parameters of the payload are known,
 or the attachment positions are fixed in advance at an appropriate position with respect to the physical parameters.

\begin{figure}[t]
    \centering
     \subfloat[Research concept]{\includegraphics[clip, width=3.0in]{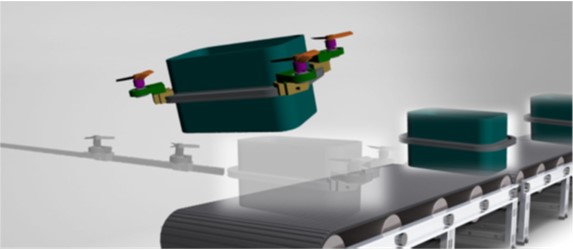}
     \label{fig:intro1}}
     \\
     \subfloat[Estimation, optimized formation, and controlled flight]{\includegraphics[clip, width=3.0in]{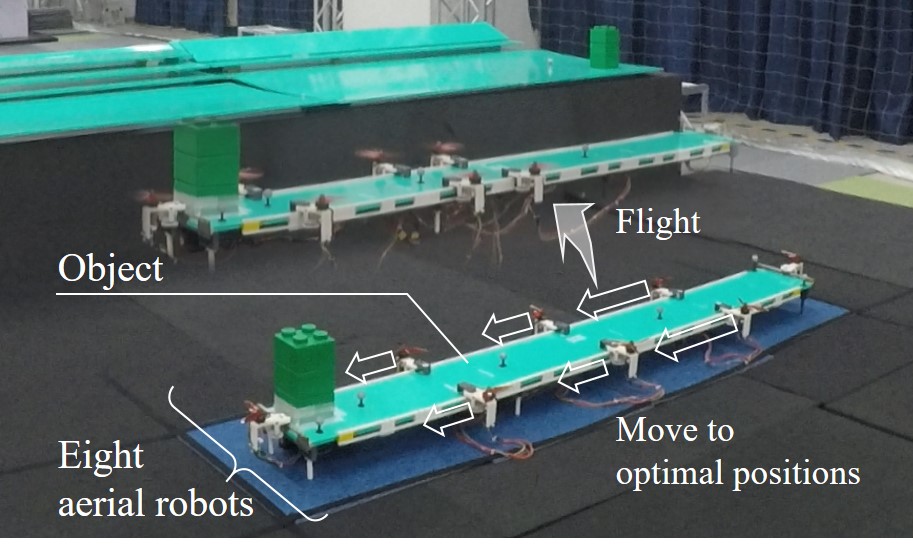}
     \label{fig:intro2}}
    
     \caption{Autonomous cooperative transportation system}
     \label{fig:intro}
\end{figure} 
Regarding the estimation of physical parameters,
extensive attempts have been made to use the interaction between ground robots and objects \cite{c10,c11,c12}.
In recent years, a method for estimating the physical parameters using aerial robots has been proposed \cite{c13,c14,c15,c15-5}.
Mellinger et al. \cite{c13} and Lee et al. \cite{c14} proposed a method of estimating physical parameters from the equation of motion under hovering conditions.
On the other hand, Corah et al. \cite{c15} proposed a method for estimating physical parameters by considering the reaction force from the ground 
and its equilibrium condition when the payload is cantilevered multiple times at a high-value attachment position based on mutual information \cite{c16}.
They adapted the interaction-based method \cite{c10,c11} and Bayesian inference method \cite{c12} used in ground-based robots for aerial robots.
They also verified the effectiveness through simulations.
This method is suitable for cooperative transportation by multiple aerial robots because it does not require to takeoff with the payload.
In a similar method, Mor\'{i}n et al. \cite{c15-5} estimated the parameters in a very short takeoff that was close to a jump.
They experimentally validated the results on lighter hardware weighing less than $100$ g.

For an aerial robot capable of autonomously estimating physical parameters,
a transportation system that can freely change the attachment position according to payload characteristics is desired.
Well-known attachment mechanisms for aerial robots include cables \cite{c4,c5,c6}, grippers with claws and magnets \cite{c8,c17}, and robot arms \cite{c14}.
Mellinger et al. \cite{c8} conducted an experiment, in which the payload was automatically gripped using a claw-shaped mechanism.
When large payloads are carried using those mechanisms, the position of the gripper being affected by the downwash or the weight of the arm poses a potential problem.
% However, the effect of downwash could not be avoided because the payload underneath was grasped by claws at the COM of the aerial robot.
% Therefore, the payload tends to be smaller than that of the aerial robots. 
% In order for a robot arm to avoid the effect of downwash, the arm needs to be extended beyond the length of the rotor, which increases the weight.

In this paper,
we introduce a novel concept of transporting unknown payloads that are larger than carrier robots,
as shown in Fig. \ref{fig:intro1}.
To realize this concept, 
we propose a cooperative transportation system that automates the entire process from pre-takeoff setting to controlled flight
using aerial robots that can move around the circumference of object.
This system can estimate the physical parameters of the object
and features a mechanism for adjusting the attachment positions according to the appropriate formation for flight, as shown in Fig.\ref{fig:intro2}.
The formation of the attachment positions is decided on site from the viewpoint of controllability.
Therefore, it is not necessary to manually investigate the characteristics of the payload and connect multiple aerial robots to appropriate positions in advance.
The contributions of this research are as follows:

\begin{itemize}
        \item We propose a novel cooperative transportation system using multiple aerial robots that can seamlessly execute the entire process of
        estimation of physical parameters, determination of optimal formation of attachment positions, rearrangement to the positions, and stable flight.
        \item It is confirmed via numerical simulations that the proposed system can transport payloads with different shapes and physical parameters.
        \item The proposed system using eight aerial robots is tested with payloads of different physical parameters.
\end{itemize}

The structure of this paper is as follows.
Sec. II describes the manufactured multiple aerial robots, the transport mechanism, and the autonomous transport system that can seamlessly execute each process.
Sec. III describes numerical experiments on two types of payloads, 
and Sec. IV describes the results of actual prototype experiments using an elongated payload. Finally, we conclude the paper in Sec. V.

\section{SYSTEM DESCRIPTION}
For multiple robots to carry unknown payloads larger than that of whole carrier robots, 
appropriate attachment positions and suitable hardware are required.
Furthermore, software executing seamlessly from decision of attachment position to flight is desirable.
As no such system has been developed thus far, we designed an architecture for such a cooperative transport system.

\subsection{Mechanical Design}
\begin{figure}[t]
        \centering
        \vspace{3mm}
         \subfloat[Mechanical design]{\includegraphics[keepaspectratio, scale=0.38]{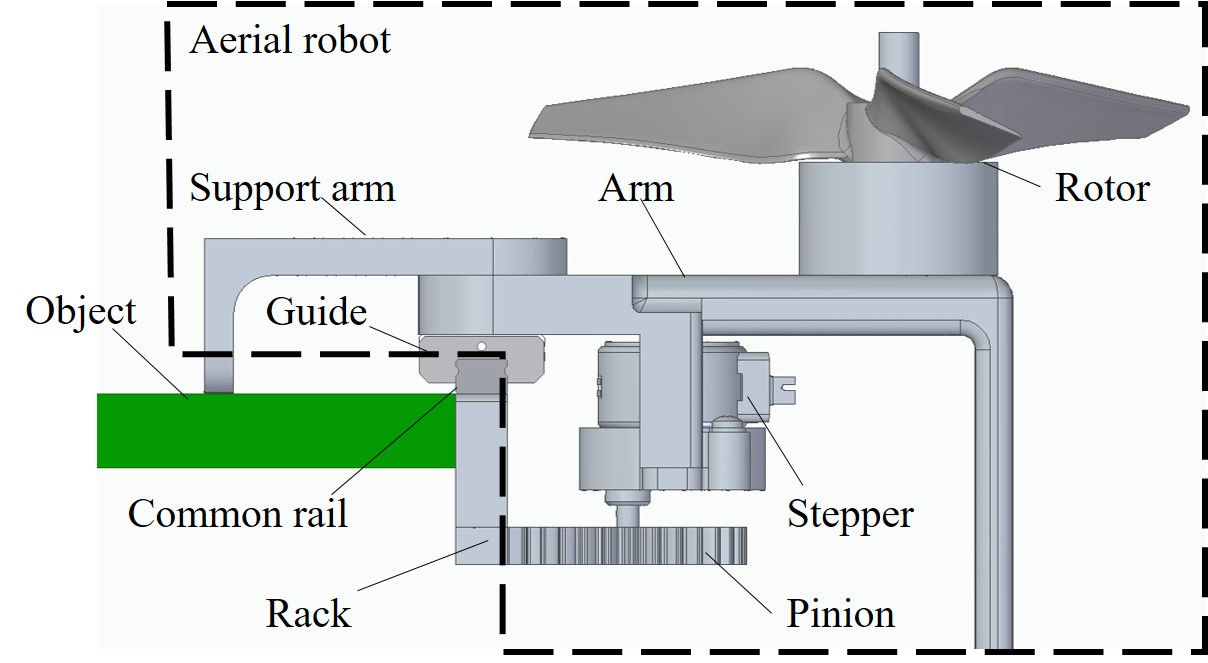}
         \label{fig:cad}}
         \\
         \subfloat[Prototype]{\includegraphics[keepaspectratio, scale=0.5]{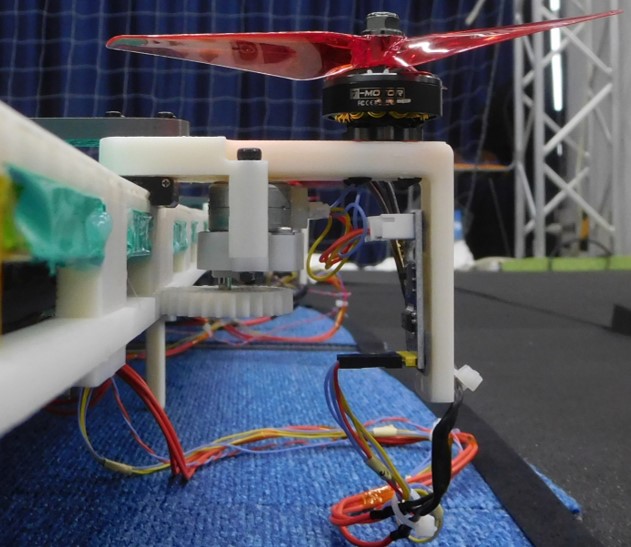}
         \label{fig:aerial}}
        
         \caption{Aerial robot moving around the payload}
         \label{fig:aerial_robot}
\end{figure}

For aerial cooperative transportation, 
the downwash generated by aerial robots strongly affects the payload.
Therefore, in our prototype, the rotors of the aerial robots are located outside the payload.
The prototype has rails attached along the shape of the payload, and the aerial robots have a variable attachment mechanism with a common rail.
As shown in Fig. \ref{fig:aerial_robot}, each aerial robot consists of a rotor, a stepping motor, and a simple arm. 
The aerial robot moves on the rail by driving the pinion gear, which meshes with the rack of the payload by the stepping motor.
In addition, a support arm improves the rigidity between the rail and the robot.
The specifications of this prototype are shown in Table \ref{table:f}.
For the thrust value, the specification value of the motor manufacturer is used.

\begin{table}[H]
        \caption{Specifications}
        \label{table:f}
        \centering
        \begin{tabular}{cc}
            \hline
            Total body size & $1.76 \times 0.2$ \; m \\
            Mass of transported object & $2.4$ \; kg \\
            Aerial robot mass & $  0.1$ \; kg \\
            60\% thrust (all robots) & $47.2$ \; N ($4.8$ \; kgf) \\
            Max thrust (all robots) & $113.6$ \; N ($11.6$ \; kgf) \\
            \hline
        \end{tabular}
\end{table}

\subsection{Software Architecture}
\begin{figure}[t]
    \centering
    \vspace{2mm}
    \includegraphics[width=86mm]{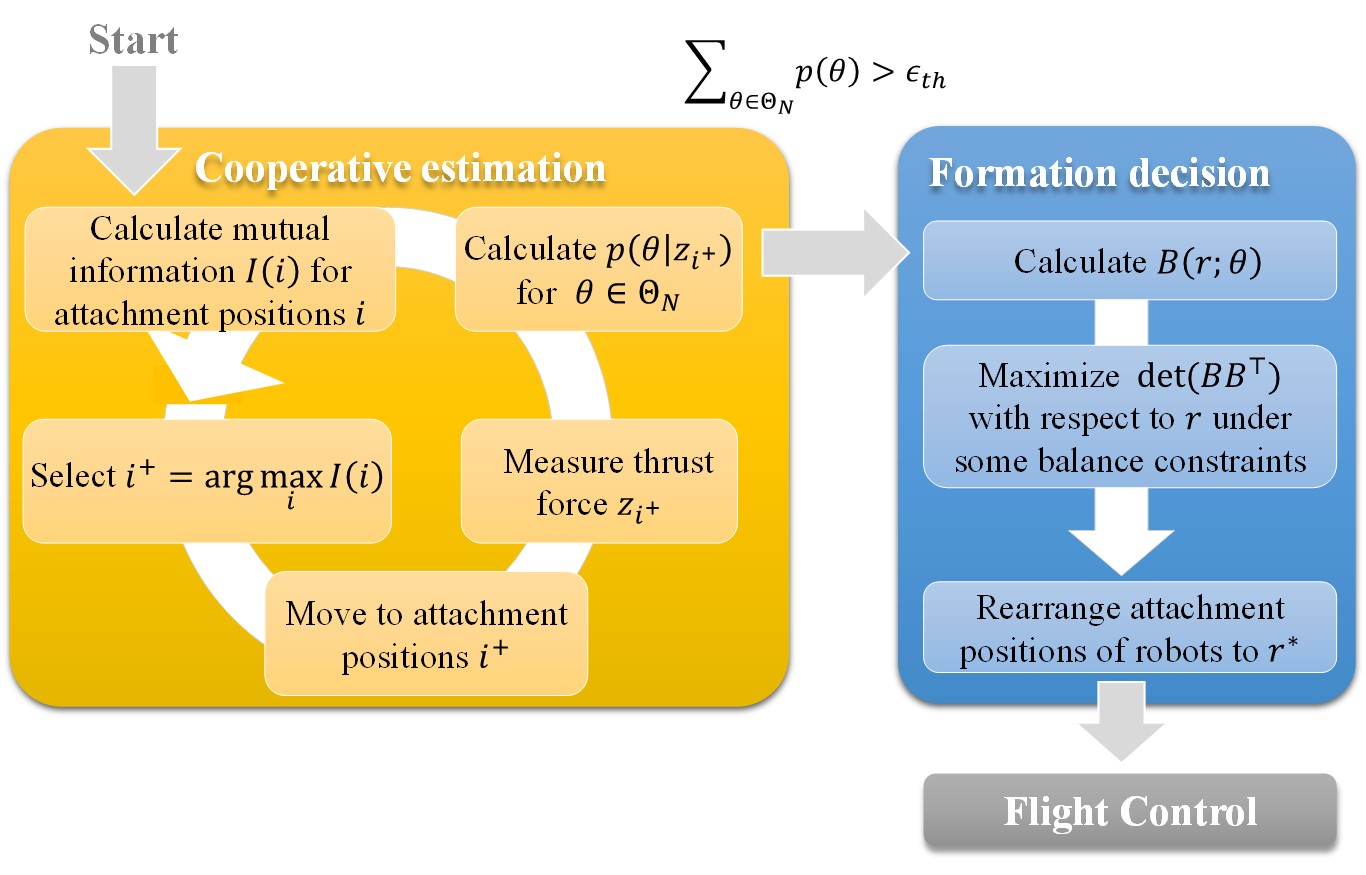}
    \caption{Software architecture from estimation to flight}
    \label{fig:arch}
\end{figure}

For transportation of an unknown payload, it is necessary to estimate the physical parameters first.
In \cite{c15}, attachment positions were simultaneously estimated and determined while increasing the number of robots using mutual information.
However, this method incurs heavy computation cost.
Therefore, 
the proposed software optimizes the formation of attachment positions from the viewpoint of controllability after estimation,  as shown in Fig. \ref{fig:arch}.
Moreover, our software seamlessly executes from estimation to flight control.
Each process is explained in the next sections.

\subsubsection{Cooperative estimation with adjacent robots}
\begin{figure}[t]
        \centering
        \vspace{3mm}
        \includegraphics[width=80mm]{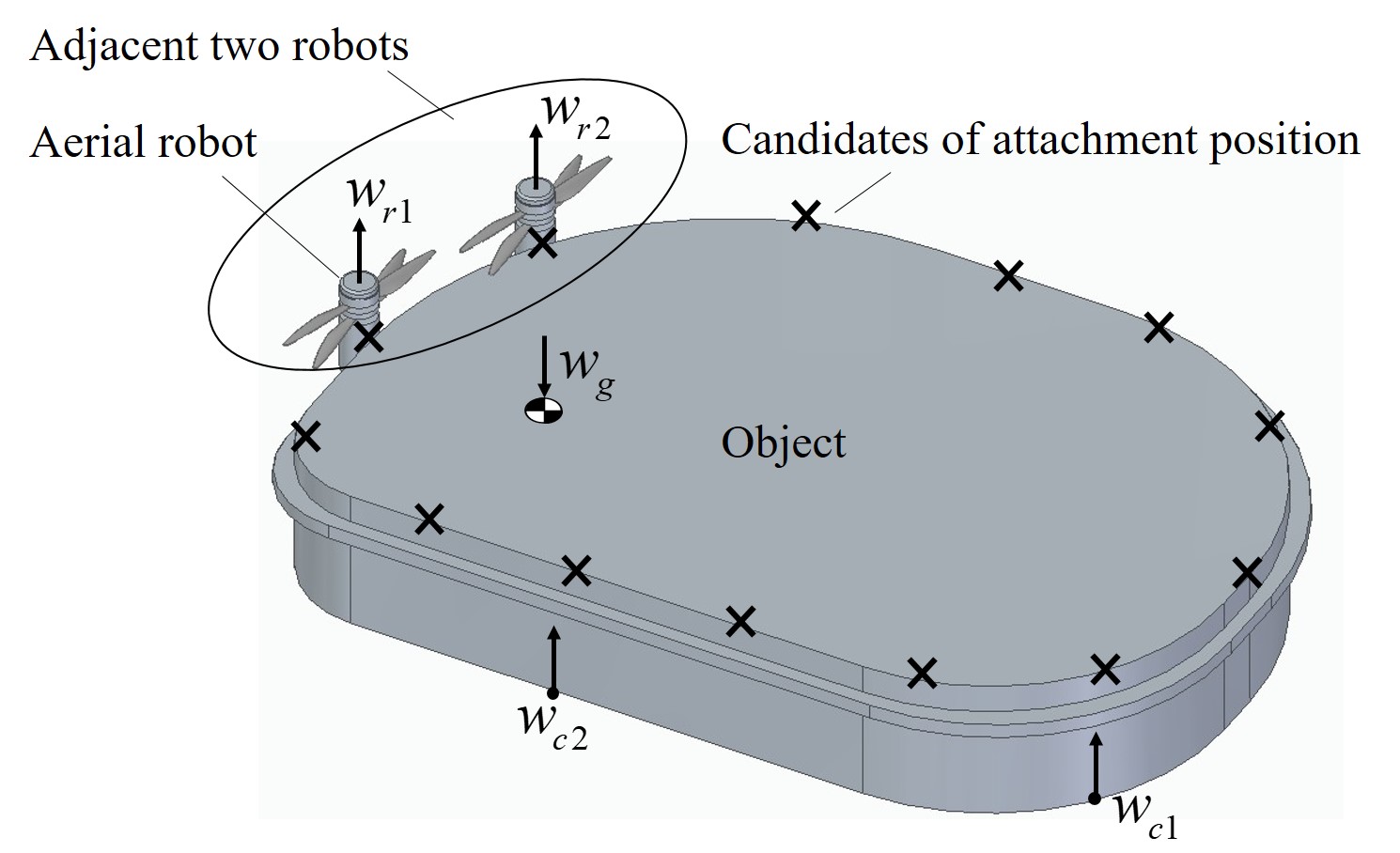}
        \caption{Adjacent paired robots during estimation process of the object}
        \label{fig:model}
    \end{figure}

In the estimation process, the shape of the payload and the number of robots required for the cantilever are assumed to be known.
For the estimation, an example with two robots is shown in Fig. \ref{fig:model}.
The estimate is made with a discrete Bayesian filter that uses the likelihood calculated physical model when the robots cantilever the payload.
The discretized physical parameters are defined as $\theta = [\theta_1,\theta_2,\theta_3]$,
where $\theta_1$ and $\theta_2$ are the COM positions and $\theta_3$ denotes the mass.
The attachment positions $i= \{i_1,i_2\}$ cantilevered by the robots are determined
using mutual information $I$ for each given candidates of attachment position $\mathcal{A}$.
To control the computational load of $I$,
$i$ is limited to the adjacent $\mathcal{A}$.
In this method, the combinations of $I$ remain the same irrespective of the number of aerial robots.

The likelihood is calculated from the physical model
when the robots apply thrust immediately after cantilevering the payload (static equilibrium condition), as shown in Fig. \ref{fig:model}.
We consider the $wrench$ vector composed of a force and torque acting on the payload, and assume that the robots have the same thrust.
In this situation, the $wrenches$ caused by the thrust of the robots, the reaction forces from the ground, and the payload are approximately in equilibrium as follows:
% Considering similar thrust by the robots $ \lambda_r$ and using $wrench \in \mathbb{R}^6 $ which represents the force and torque vector,
% the static equilibrium equation is as follows:
\vspace{-0.5mm}
\begin{align}
    \label{eq:fr}
    \begin{split}
        & W_C \Lambda _C + w_g + {W_0 ^i} + {W_r ^i} {\lambda _r} = 0,\\
        \vspace{3mm}
        &s.t. \hspace{5.8mm} {\lambda _r} \ge 0, \Lambda _C \ge 0, \\
        &\hspace{10mm} W_0^i = w_0^{i_1} + w_0^{i_2}, \\
        &\hspace{10mm} W_r^i = w_r^{i_1} + w_r^{i_2},
    \end{split}
\end{align}
where $C$ is a set of contact points $c$, $W_C$ is a matrix with some of the normalized $wrenches \; w_c \in \mathbb{R}^r \; (c \in C)$. 
$\Lambda_C$ is a vector with reaction force $\lambda_c$, and $w_g$ is $wrench$ at the COM, 
$w_0^{i_1}$ and $w_0^{i_2}$ are $wrench$ of the robot's own weight, 
$w_r^{i_1}$ and $w_r^{i_2}$  are the normalized $wrenches$ applied by the robot, and $ \lambda_r$ is thrust by the robots.
Given the measurement $z_i$ of $\lambda _r$ satisfying equation (\ref{eq:fr}) as a Gaussian distribution, the likelihood $ p(z_i | \theta)$ is as follows:
\vspace{-0.25mm}
\begin{equation}
    \label{eq:yudo}
    p(z_i | \theta)= \textit{N}(z_i ; {\lambda _r}(\theta),\sigma ^2) = \textit{N} _{\theta}(z_i),
\end{equation}
where $ \textit{N}(a;\mu , \sigma ^2)$ denotes a normal distribution in which the random variable $a$ has a mean $\mu$ and variance $\sigma^2$.

The flow of estimation using this likelihood is shown on the left in Fig. \ref{fig:arch}.
First, $I$, which indicates the value of the paired attachment positions, is calculated using the method described in \cite{c15},
using the probability $p(\theta)$ (the initial value is uniform) and likelihood. 
Second, the $i^*$ with the maximum $I$ is selected.
Third, the paired aerial robots are moved to $i^*$.
Fourth, $z_{i^+}$ is measured when the transported object is cantilevered.
Fifth, using (\ref{eq:yudo}) and the Bayes' theorem,
\vspace{-0.25mm}
\begin{equation*}
    \label{eq:beiz}
    p(\theta | z_i) = \frac{p(z_i | \theta)p(\theta)}{\sum_{\theta \in \Theta} p(z_i | \theta)p(\theta)},
\end{equation*}
the posterior $p(\theta | z_i)$ is determined.
Finally,  $p(\theta)$ is updated by $p(\theta | z_i)$.
In our method, $\Theta_N$ is a set of points within resolution of 1 from $\theta$ with the highest $p(\theta)$,
and the estimation continues until $\Sigma_{\theta \in \Theta_N} p(\theta)$  exceeds the threshold $\epsilon_{th}$.

\subsubsection{Formation decision}
The attachment positions for flight are determined based on the estimated physical parameters.
In flight, an even thrust load ratio and high controllability during hovering are desirable.
Therefore, we set a balance constraint when all robots apply the same thrust for the roll and pitch, 
and determine the attachment positions to ensure the maximum controllability Gramian.
With the estimated COM position as the origin of the body coordinate system, 
the state equation of our hardware, including the aerial robot, related to roll, pitch, and yaw is as follows:
\vspace{-0.25mm}
\begin{eqnarray}
    \label{eq:AB}
      \dot{\zeta} = A\zeta + Bu,
\end{eqnarray}
where $\zeta \in \mathbb{R}^3$ is the state of roll, pitch, and yaw angles in the body coordinate system, 
$u \in \mathbb{R}^{n_r}$ is the input vector, $n_r$ is the number of robots,
$A \in \mathbb{R}^{3 \times 3}$ is the state matrix, 
and $B \in \mathbb{R}^{3 \times n_r}$ is the input matrix determined by the attachment positions of the robots and the COM of the payload.
The controllability Gramian in equation (\ref{eq:AB}) is defined by
\vspace{-0.25mm}
\begin{eqnarray*}
    \label{eq:gram}
    \int^{\infty}_0 e^{At}BB^{\top}e^{A^{\top}t}dt.
\end{eqnarray*}
When $A$ is approximately zero, the maximization of the controllability Gramian is equivalent to that of $BB^\top$.
Using the attachment position $r^j = [r^j_1, r^j_2]^\top$ of $j$-th robot in the body coordinate system,
the thrust coefficient $c_t$, and the torque coefficient $c_q$,
$B$ is as follows:
\begin{equation*}
    \label{eq:B}
    B = \left(
        \begin{array}{cccc}
          c_tr^1_1 & c_tr^2_1 & \ldots  & c_tr^{n_r}_1 \\
          c_tr^1_2 & c_tr^2_2 & \ldots  & c_tr^{n_r}_2 \\
          s_1c_q & s_2c_q & \ldots  & s_{n_r}c_q
        \end{array}
      \right)
      = \left(
      \begin{array}{c}
        b_1^{\top} \\
        b_2^{\top} \\
        b_3^{\top}
      \end{array}
      \right),
\end{equation*}
where $[s_1,s_2,\cdots,s_{n_r}]$ are signs denoting the direction of rotation of the rotor,
and $b_1, b_2, b_3 \in \mathbb{R}^3$ are the roll, pitch, and yaw elements of matrix $B$, respectively.
Therefore, the optimal attachment positions $r^\ast$ that satisfies the following condition are desirable:
\vspace{-0.5mm}
\begin{equation}
    \label{eq:J}
    \begin{split}
    & r^\ast =\arg\max_r det(BB^T),\\
    \vspace{3mm}
    &s.t. \hspace{8.2mm}  \left|\sum_{k=1}^{n_r} b_{1}(k)\right| \leq \epsilon, \\
    &\hspace{13mm} \left|\sum_{k=1}^{n_r} b_{2}(k)\right| \leq \epsilon,
    \end{split}
\end{equation}
where $r=[{r^1}^{\top}, ..., {r^{n_r}}^{\top}]^{\top}$ and $\epsilon$ is a sufficiently small value.
Inequalities of (\ref{eq:J}) are added to prevent the robots from losing their balance when the same thrust is applied during takeoff.
The aerial robots are rearranged to the determined $r^*$.

\subsubsection{Flight control}
For flight control, general PID control, similar to that in literature \cite{c15-5}, is used.
The controller is a cascaded PID with four PID controllers: position, velocity, angle, and angular velocity.
The mixing matrix $B^{\top}$ is recalculated using $r^*$.
In addition, the estimated mass is used in the gravity compensation.

\begin{figure}[b!]
    \begin{minipage}[b]{0.5\linewidth}
        \centering
        \includegraphics[keepaspectratio, scale=0.32]{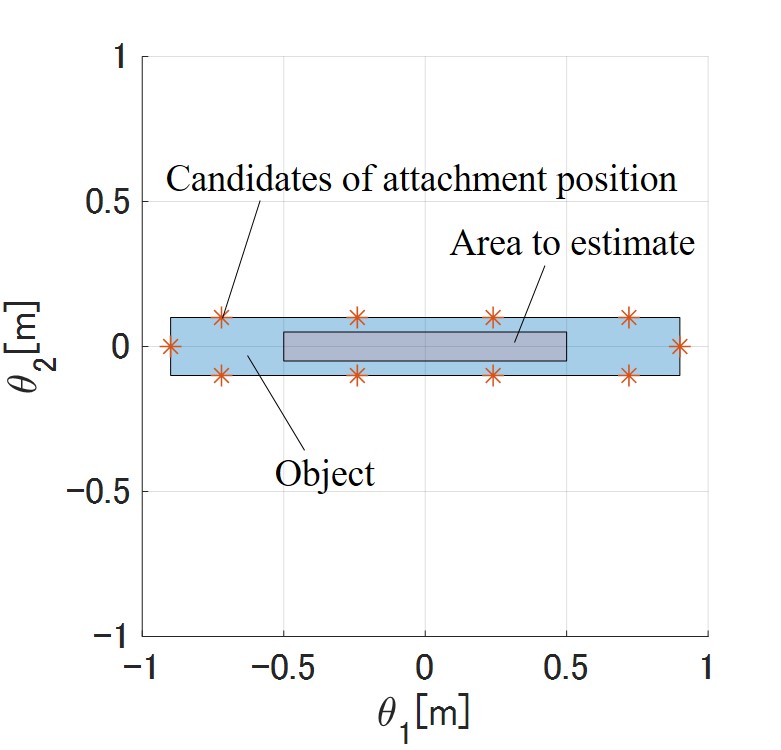}
        \subcaption{Rectangle-shape}\label{fig:ita_con}
    \end{minipage}
    \begin{minipage}[b]{0.49\linewidth}
        \centering
        \includegraphics[keepaspectratio, scale=0.32]{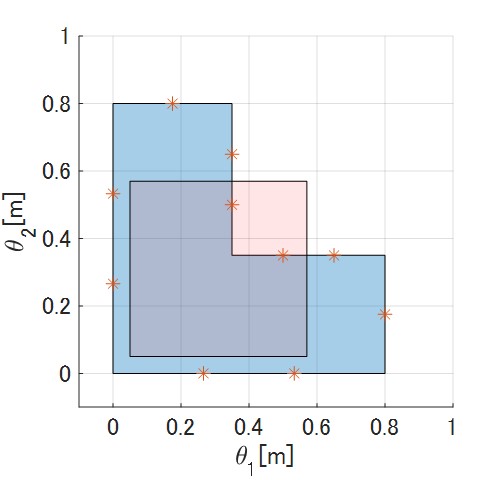}
        \subcaption{L-shape}\label{fig:L_con}
    \end{minipage}
    \caption{Two different shapes for simulation.
    The blue area is the shape of the object. 
    The red area is the range where the COM can be taken. 
    The red asterisks are candidates of attachment position.
    }\label{fig:sim}
\end{figure}

\begin{figure*}[b]
    \begin{minipage}[b]{0.33\linewidth}
        \centering
        \includegraphics[keepaspectratio, scale=0.33]{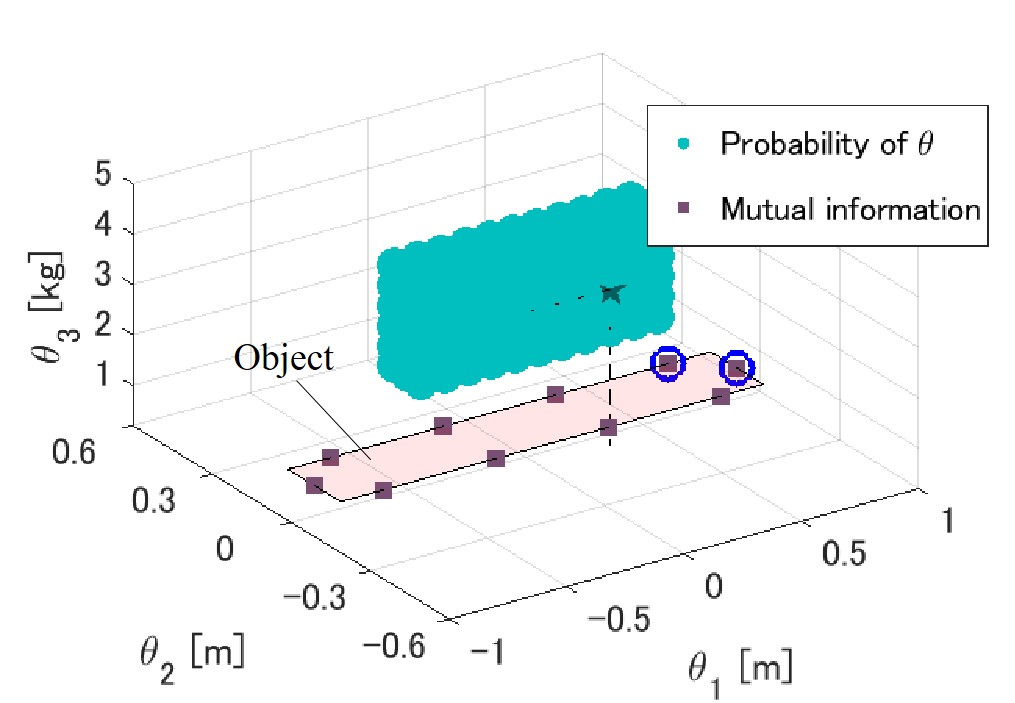}
        \subcaption{Initial state}\label{fig:p_sim1}
    \end{minipage}
    \begin{minipage}[b]{0.33\linewidth}
        \centering
        \includegraphics[keepaspectratio, scale=0.33]{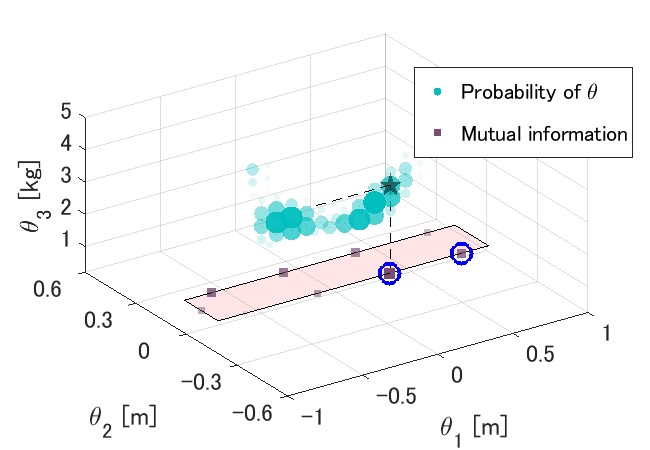}
        \subcaption{2nd measurement}\label{fig:p_sim2}
    \end{minipage}
    \begin{minipage}[b]{0.33\linewidth}
        \centering
        \includegraphics[keepaspectratio, scale=0.33]{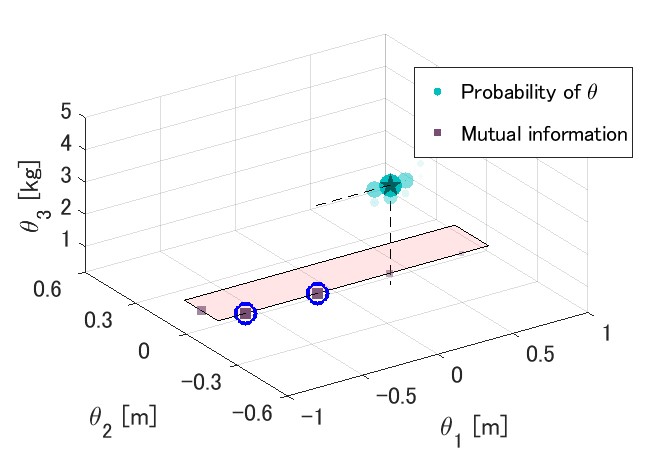}
        \subcaption{4th measurement}\label{fig:p_sim3}
    \end{minipage}\\
    \begin{minipage}[b]{0.33\linewidth}
            \centering
            \includegraphics[keepaspectratio, scale=0.33]{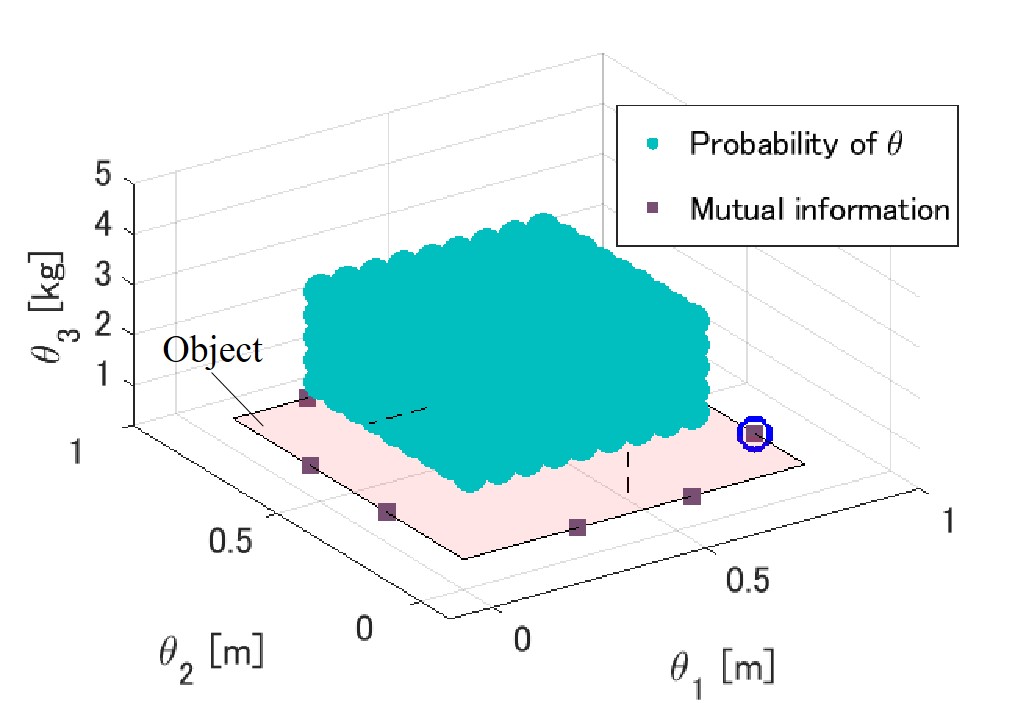}
            \subcaption{Initial state}\label{fig:L_sim1}
        \end{minipage}
        \begin{minipage}[b]{0.33\linewidth}
            \centering
            \includegraphics[keepaspectratio, scale=0.33]{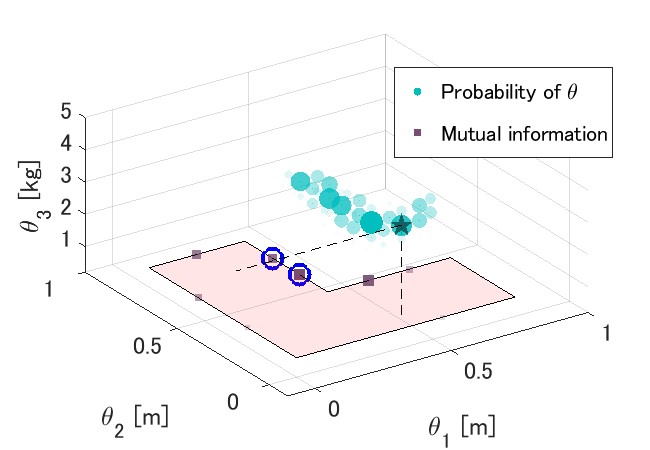}
            \subcaption{2nd measurement}\label{fig:L_sim2}
        \end{minipage}
        \begin{minipage}[b]{0.33\linewidth}
            \centering
            \includegraphics[keepaspectratio, scale=0.33]{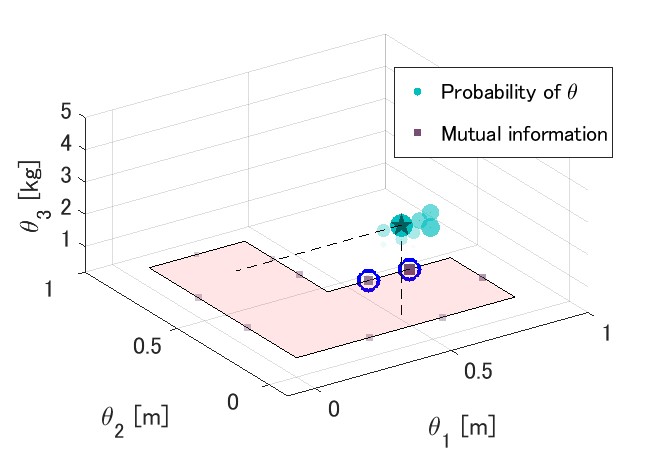}
            \subcaption{4th measurement}\label{fig:L_sim3}
        \end{minipage}
    \caption{Physical parameter estimation for two shapes.
    The upper row shows the rectangle and the lower row shows the L-shape. 
    The size of the sphere in the center of each graph indicate the probability of $\theta$,
    and the sphere position indicates the position and mass of the COM. 
    The star denotes the true value.
    The square at the bottom of the graph shows the candidates of attachment position, 
    and the size of the square indicates the difference in the amount of mutual information.
    The blue circle shows the pair of attachment positions to be measured next.}\label{fig:sim_r}
\end{figure*}

\begin{figure}[h]
    \centering
    \vspace{3mm}
    \includegraphics[keepaspectratio, scale=0.45]{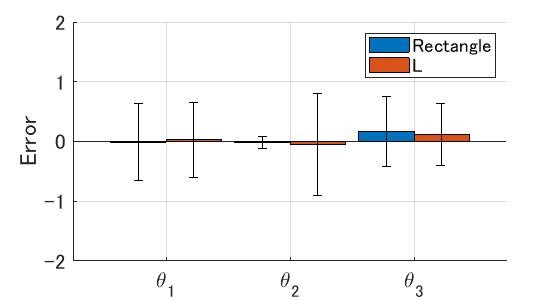}
    \caption{Error during simulations with 100 different COMs using two shapes. 
    The errors are normalized by the resolution.
    The bar graphs show the average values, and the error bars show the standard deviation.}\label{fig:sim_error}
\end{figure}

\begin{figure}[h]
    \centering
    \subfloat[Rectangle-shape]{\includegraphics[keepaspectratio, scale=0.38]{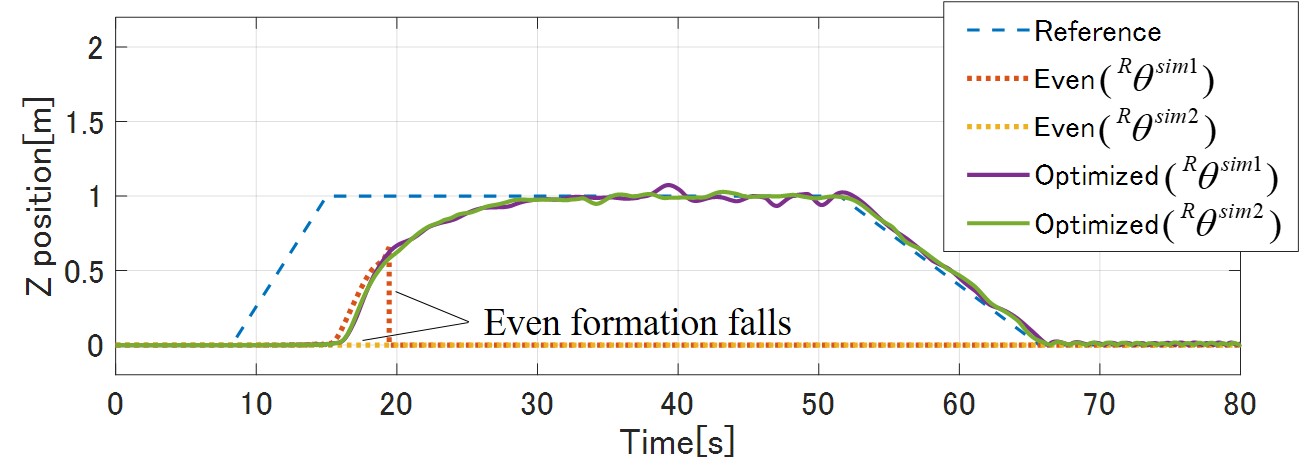}
    \label{fig:sim_rec_fall}}
    \\
    \subfloat[L-shape]{\includegraphics[keepaspectratio, scale=0.38]{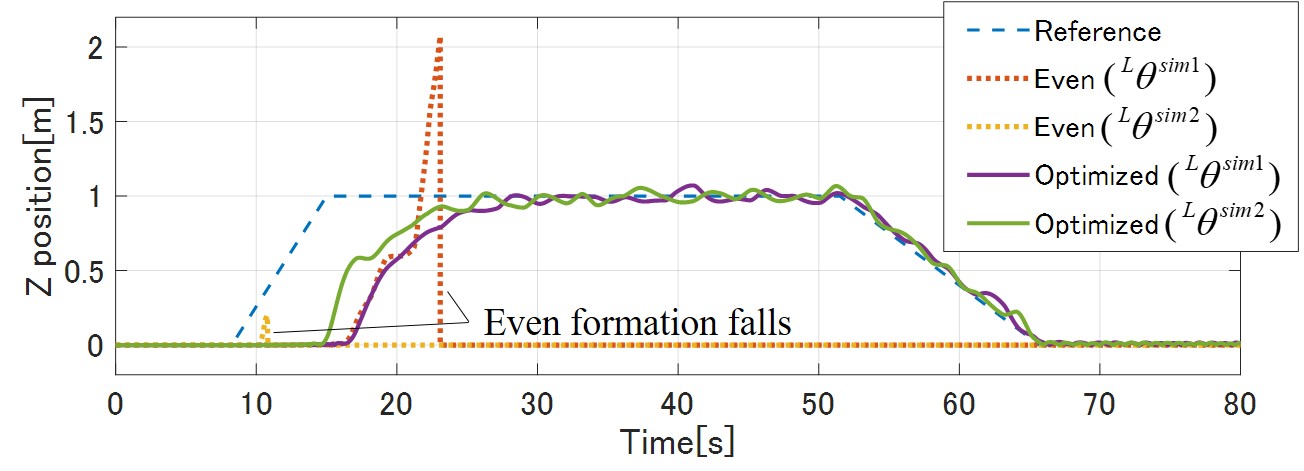}
    \label{fig:sim_L_fall}}
   
    \caption{Z position during flight under the optimized formation and even formation.
    The dashed line is the reference value, the dotted line is the actual value for the even formation, and the solid line is the actual value for the optimized formation.
    }\label{fig:sim_fly}
\end{figure}

\section{Simulation}
Two types of simulations with eight robots were performed, one using the rectangle-shaped prototype and the other using the L-shaped object. 
First, to confirm the estimation, physical parameters were randomly set and the estimation process was performed 100 times.
Next, to confirm the series of functions up to flight, two physical parameters were set and the object was transported to a target position.
All these processes were programmed in MATLAB Simulink and Stateflow.

\subsection{Condition}
Fig. \ref{fig:sim} shows two shapes for simulations.
For the rectangle-shape in the left graph,
the estimation ranges of $\theta (= [\theta_1, \theta_2, \theta_3])$ were $-0.5$ to $0.5$ m for $\theta_1$, 
$-0.05$ to $0.05$ m for $\theta_2$, and $2.5$ to $4$ kg for $\theta_3$.
The resolutions of $\theta$ were $[0.1,0.03,0.5]$.
For the L-shape in the right graph,
the estimation ranges of $\theta$ were $0.05$ to $0.57$ m for $\theta_1$, $0.02$ to $0.57$ m for $\theta_2$, and $2.2$ to $4$ kg for $\theta_3$.
The resolutions of $\theta$ were $[0.065,0.065,0.45]$.
The estimated threshold $\epsilon_{th}$ is $0.8$ for both simulations.
In the simulations, the measured thrust force was $\lambda_r$ in (\ref{eq:fr}) with white noise (mean $0$ and variance $1$).
The true values of physical parameters were randomly changed 100 times.

Subsequently, the optimal formation calculated by (\ref{eq:J}) was compared with the even formation,
and two true physical parameters were set for each shape:
$ ^{R} \theta^{sim1} = [0.33, 0.01, 3.5]$, $ ^{R} \theta^{sim2} = [0.4, 0.01, 3]$, 
$ ^{L} \theta^{sim1} = [0.31, 0.1, 3.5]$, and $ ^{L} \theta^{sim2} = [0.4, 0.1, 3.5]$.
The simplex search method \cite{c19} was used for the optimization, with $\epsilon = 10^{-5}$ for the rectangle-shape and $\epsilon = 0.1$ for the L-shape.
In the flight simulation,
the reference trajectory was given in three dimensions, and the disturbance of the body torque was added to (\ref{eq:AB}).
If the roll or pitch angle of the object exceeded $45^\circ$, the object was assumed to have been dropped.

\subsection{Result}
An example of a physical parameter estimation is presented in Fig. \ref{fig:sim_r}.
It can be confirmed that the probability of $\theta$ gradually converges from uniform distribution to the true value.
Fig. \ref{fig:sim_error} shows the average and standard deviation of physical parameter errors in 100 simulations.
Our method succeeded in estimating the standard deviation within resolution of 1.

The results of flight simulation for two physical parameters are shown in Fig. \ref{fig:sim_fly}.
If the physical parameters are out of alignment and the formation is uniform, 
uneven torque is generated in the roll and pitch during takeoff, causing large attitude fluctuations.
In addition, even if takeoff is possible, 
the robots are vulnerable to disturbance as high power is required for balancing.
By optimizing the formation, our system minimized imbalances and stabilized takeoff and flight.

\begin{figure*}[t]
    \vspace{2mm}
    \begin{minipage}[b]{0.32\linewidth}
        \centering
        \includegraphics[keepaspectratio, scale=0.34]{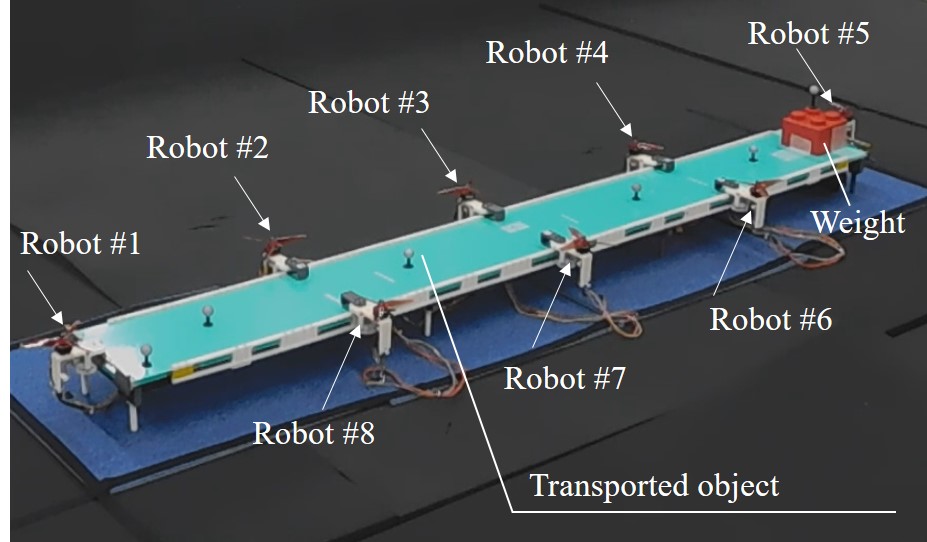}
        \subcaption{Initial formation}\label{fig:mv1}
    \end{minipage}
    \begin{minipage}[b]{0.35\linewidth}
        \centering
        \includegraphics[keepaspectratio, scale=0.34]{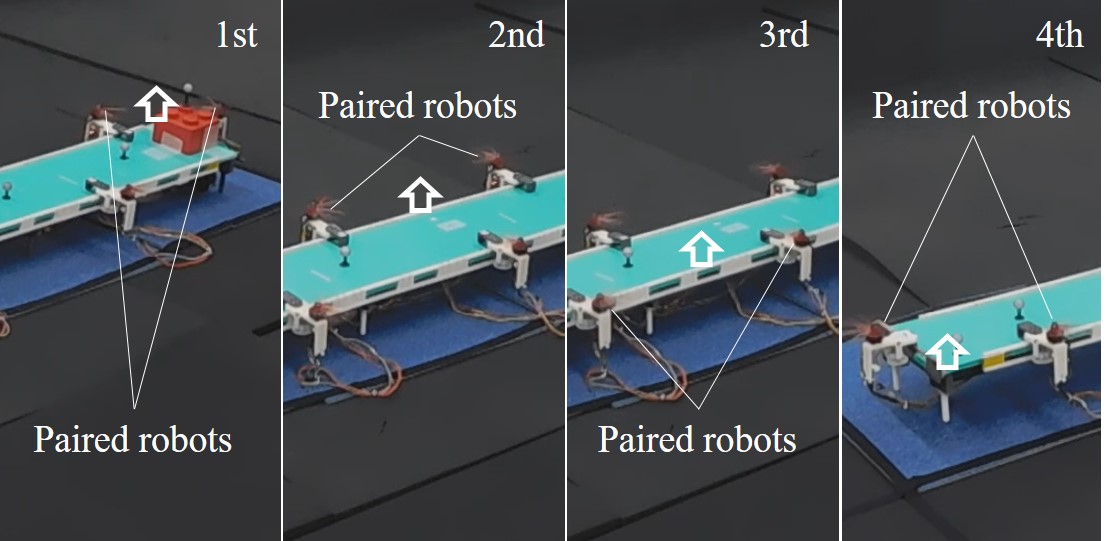}
        \subcaption{Parameter estimation}\label{fig:mv2}
    \end{minipage}
    \begin{minipage}[b]{0.33\linewidth}
        \centering
        \includegraphics[keepaspectratio, scale=0.34]{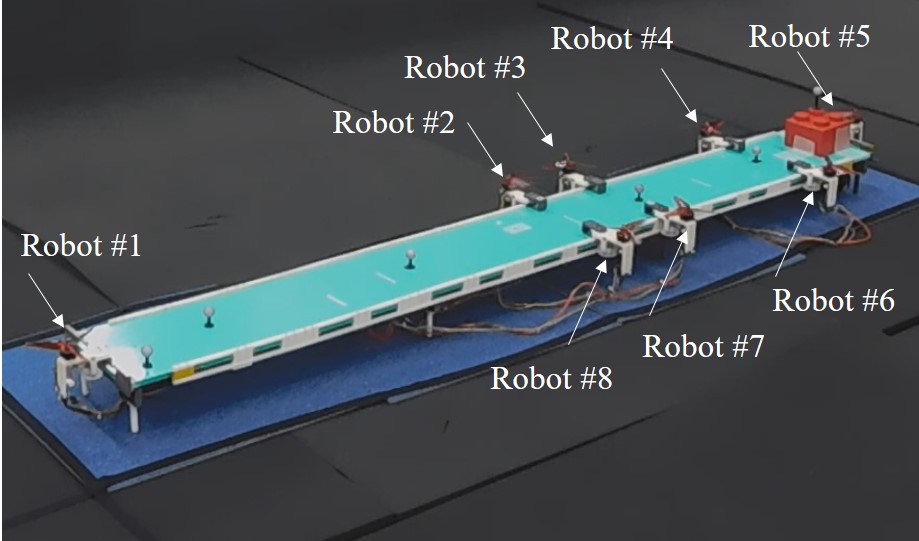}
        \subcaption{Optimized formation}\label{fig:mv3}
    \end{minipage}
    \caption{Sequential shot of estimation and formation decision under $\theta^{ex1}$}\label{fig:mv}
\end{figure*}

\begin{figure}[!t]
    \centering
    \includegraphics[keepaspectratio, scale=0.45]{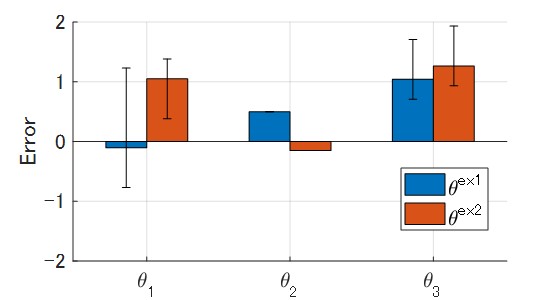}
    \caption{Experiment results of estimation error with $\theta^{ex1}$ and $\theta^{ex2}$.
             The errors are normalized by the resolution. The bar graph shows the average, and the error bar shows the maximum/minimum value.
             No error bar is provided for $\theta_2$ because all the estimates are the same.
            }\label{fig:ex_error}
    \vspace{5mm}

    \centering
    \subfloat[Even formation]{\includegraphics[clip, width=2.3in]{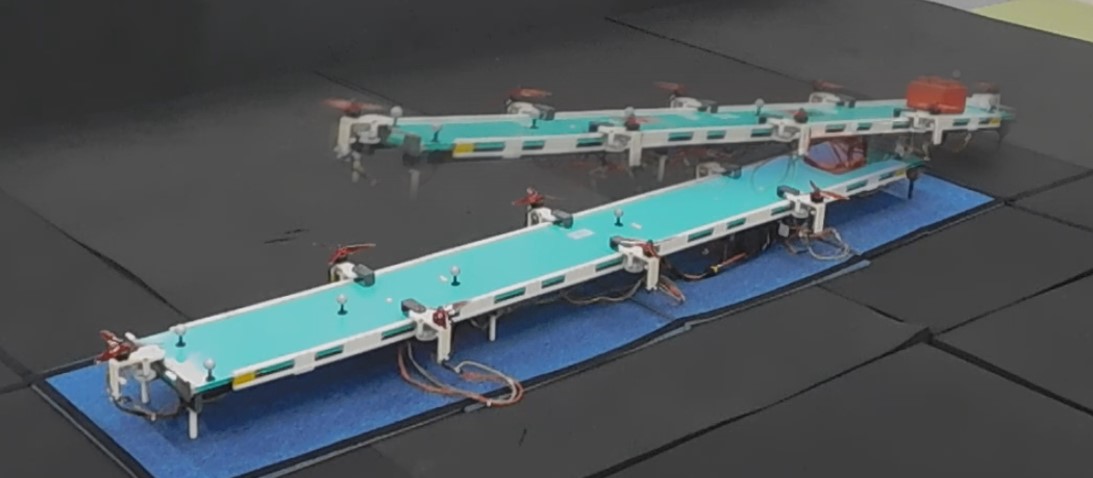}
    \label{fig:mv_fall}}
    \\
    \subfloat[Optimized formation by proposed system]{\includegraphics[clip, width=2.3in]{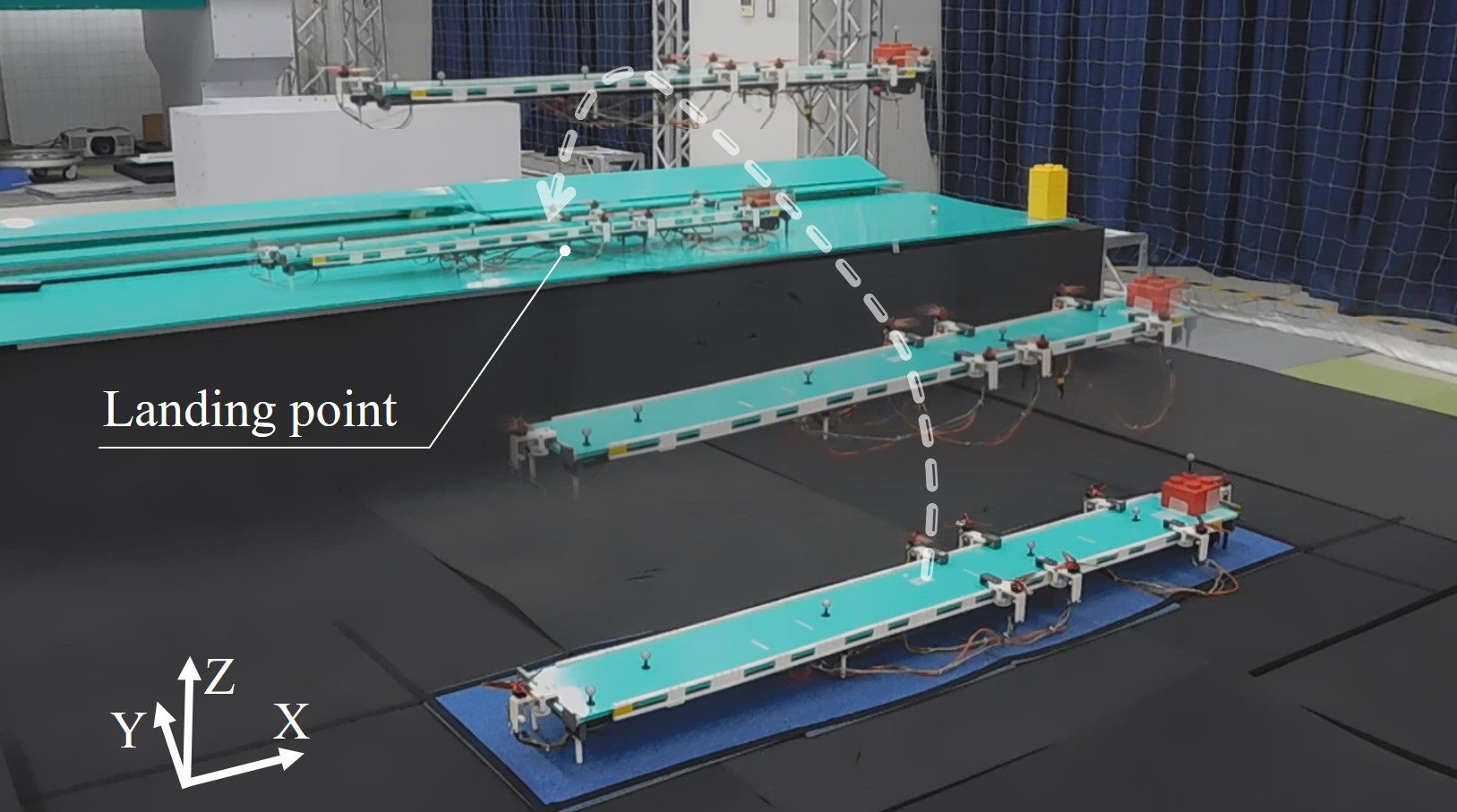}
    \label{fig:mv_fly}}
    \caption{Flight demonstration}
    \label{fig:ex_fly}
 \end{figure}
 
%  \begin{figure}
%     \centering
%      \subfloat[Even formation]{\includegraphics[clip, width=2.3in]{ex_fall.jpg}
%      \label{fig:mv_fall}}
%      \\
%      \subfloat[Proposed system]{\includegraphics[clip, width=2.3in]{ex_pic_fly.jpg}
%      \label{fig:mv_fly}}
%      \caption{Flight experiment}
%      \label{fig:ex_fly}
% \end{figure}

\begin{figure}[!t]
    \centering
    \begin{minipage}[b]{0.4\linewidth}
        \centering
        \includegraphics[keepaspectratio, scale=0.31]{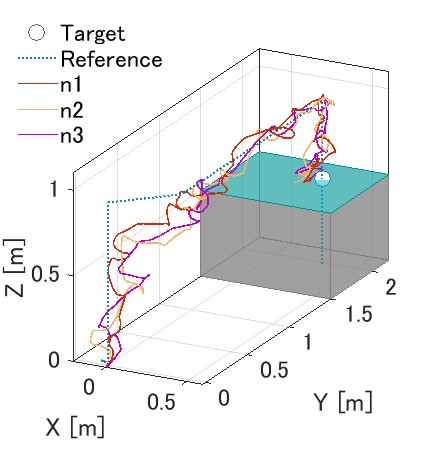}
        \subcaption{Trajectory}\label{fig:ex1_3D}
    \end{minipage}
    \begin{minipage}[b]{0.5\linewidth}
        \centering
        \includegraphics[keepaspectratio, scale=0.31]{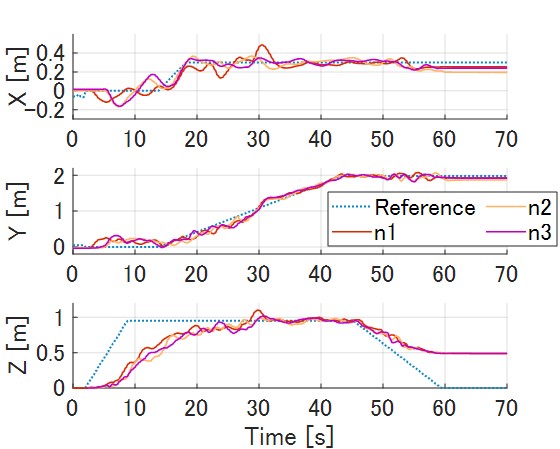}
        \subcaption{Time series of position}\label{fig:ex1_time}
    \end{minipage}\\
    \begin{minipage}[b]{0.4\linewidth}
        \centering
        \includegraphics[keepaspectratio, scale=0.31]{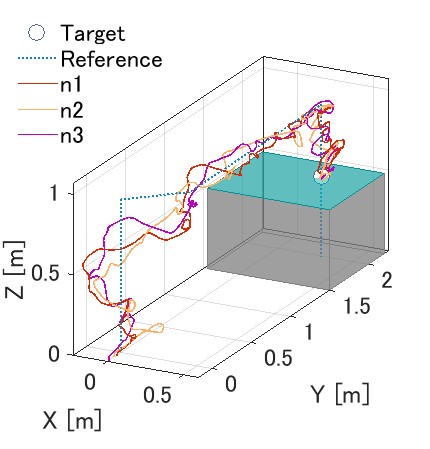}
        \subcaption{Trajectory}\label{fig:ex2_3D}
    \end{minipage}
    \begin{minipage}[b]{0.5\linewidth}
        \centering
        \includegraphics[keepaspectratio, scale=0.31]{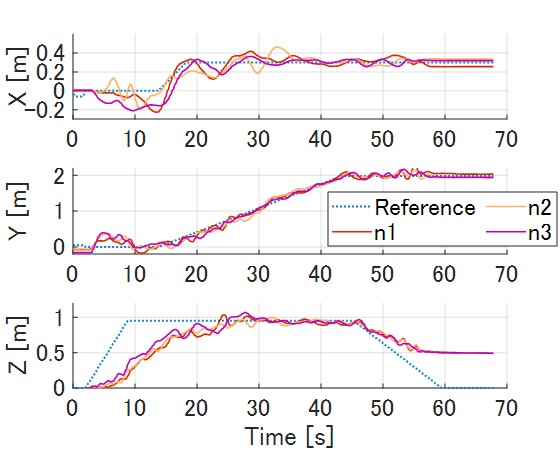}
        \subcaption{Time series of position}\label{fig:ex2_time}
    \end{minipage}
    \caption{Positions during three flight experiments. 
    (a) and (b) correspond to $\theta^{ex1}$ and (c) and (d) correspond to $\theta^{ex2}$.
    The dotted lines represent reference trajectory, and the solid lines represent the actual positions of three times.
    Reference $Z = 0$ represents a landing command.
    }\label{fig:ex_result}
\end{figure}

\section{Experiment}
We tested the designed prototype described in Sec. II.
The software was created by MATLAB Simulink and Stateflow as well as the simulation and implemented in the controller PIXHAWK \cite{c20}.
The batteries were mounted on the payload side and were shared between multiple aerial robots.
In fact, transportation of objects with a weight of approximately $3.2$ kg (with additional weight) and a maximum length of $1.76$ m
by small aerial robots that are unaware of the physical parameters of the payload has rarely been demonstrated.

\subsection{Condition}
In the estimation, it is necessary to detect the equilibrium state and measure the thrust.
We detected the equilibrium state from the change in the attitude value calculated from the inertial measurement unit (IMU) in the controller.
In addition, the thrust was derived based on the command-thrust map published by the manufacturer because it cannot be measured directly.
Furthermore,
to exclude the weight of robots from payload weight, 
unselected robots applied only the thrust of their own weight during the measurement.

In the optimization of formation with $\epsilon = 0$,
from the practical viewpoint,
the four attachment positions were decided by the estimated COM, the other positions were optimized by the symmetrical arrangement constraint as well as (\ref{eq:J}).

The experiment was performed three times each under the conditions $\theta^{ex1} = [0.21,0,3.23]$
and $\theta^{ex2}= [-0.17,0.014,3.13]$.
These were set using additional weight.
The estimation ranges of $\theta$ were $-0.26$ to $0.26$ m for $\theta_1$, $-0.06$ to $0.06$ m for $\theta_2$, and $2.2$ to $4$ kg for $\theta_3$.
The resolutions of $\theta$ were $[0.065,0.04,0.45]$ and the estimated threshold $\epsilon_{th}$ is $0.53$.
The $X$, $Y$, and $Z$ positions of the target were set to $0.3$, $1.98$, and $0.5$, respectively.
The reference trajectory from the initial position to the target position was used by the flight controller.
The position of the object was measured by a motion capture system, and the attitude was calculated by IMU in the controller. 

\subsection{Result}
Fig. \ref{fig:mv} shows photographic sequence from estimation to decision of the formation of the attachment positions.
As can be seen in Fig. \ref{fig:mv}, the formation was optimized to focus thrust on the right side due to the additional weight.
Fig. \ref{fig:ex_error} shows 
that the estimation was performed properly because the estimation error of the COM position was close to a resolution of 1.
However, the mass error was biased in the positive direction.
This is because the thrust is measured as larger than its actual value because of the error in the command-thrust map 
and the delay in detection due to backlash and deflection.

Fig. \ref{fig:ex_fly} illustrates the results of the aerial transportation with $\theta^{ex1}$.
In the case of even formation, eight robots could not take off, as shown in Fig. \ref{fig:mv_fall}.
When the formation was optimized by the proposed method, successful transportation to the target position could be achieved
against the payload with the COM shifted to the right side, as shown in Fig. \ref{fig:mv_fly}.
In all three experiments, our system succeeded in takeoff and tracking the payload to the reference, as shown in Figs. \ref{fig:ex1_3D} and \ref{fig:ex1_time}.
In addition, our system successfully transported the payload with different physical parameters, as shown in Figs. \ref{fig:ex2_3D} and \ref{fig:ex2_time}.

\section{CONCLUSIONS}
In this paper, 
we proposed an aerial cooperative transportation system for unknown payloads larger than that of whole carrier robots.
The system comprised transport hardware with rails on which the aerial robot could move
and a controller that seamlessly executed the process from preparation before takeoff to flight.
Using the manufactured prototype, 
we successfully performed aerial transport of unknown payloads (with a weight of about $3.2$ kg and maximum length of $1.76$ m) with off-center COMs using eight robots.
Thus, limitations related to the payload can be relaxed, and the cooperative transportation is safer and more autonomous.
In the future, 
we plan to optimize the formation to adapt against unexpected shift of the load during flight and relax the hardware restrictions by decentralizing control of the aerial robots.


%%%%%%%%%%%%%%%%%%%%%%%%%%%%%%%%%%%%%%%%%%%%%%%%%%%%%%%%%%%%%%%%%%%%%%%%%%%%%%%%

%%%%%%%%%%%%%%%%%%%%%%%%%%%%%%%%%%%%%%%%%%%%%%%%%%%%%%%%%%%%%%%%%%%%%%%%%%%%%%%%



% \section*{ACKNOWLEDGMENT}

% %%%%%%%%%%%%%%%%%%%%%%%%%%%%%%%%%%%%%%%%%%%%%%%%%%%%%%%%%%%%%%%%%%%%%%%%%%%%%%%%

% References are important to the reader; therefore, each citation must be complete and correct. If at all possible, references should be commonly available publications.

\end{document}